\documentclass[11pt]{article}

\usepackage[final]{acl}

\usepackage{times}
\usepackage{latexsym}
\usepackage[T1]{fontenc}
\usepackage[utf8]{inputenc}
\usepackage{microtype}
\usepackage{inconsolata}
\usepackage{graphicx}
\usepackage{booktabs}
\usepackage{multirow}
\usepackage{amsmath}
\usepackage{amssymb}
\usepackage{csquotes}
\usepackage{seqsplit}
\usepackage{array}
\usepackage{pgfplots}
\pgfplotsset{compat=1.18}
\usepackage{hyperref}

\title{Right Tool, Right Job: Native-Language Evaluation, Tokenizer Sensitivity, and Methodological Findings from a French-Only BabyLM}

\author{
Adam Zachary Wasserman$^{1}$ \quad David Beauchemin$^{2}$ \\
$^{1}$Open Honest Foundation \\
$^{2}$Group for Research in Artificial Intelligence of Laval University (GRAIL), \\
Universit\'e Laval, Qu\'ebec, Canada \\
\texttt{adam.wasserman@openhonest.org} \quad \texttt{david.beauchemin@ift.ulaval.ca}
}

\begin{document}
\maketitle

\begin{abstract}
We submit MéTRON-FR, a 125M GPT-2 pretrained on 92.47M words of French, to the BabyLM 2026 Strict track. It scores $85.97 \pm 0.17$\% on QFrBLiMP (a native Quebec-French benchmark of grammatical minimal pairs) and $62.80$\% on the BabyLM-weighted leaderboard.
A cross-lingual GLUE (General Language Understanding Evaluation) protocol that combines French task-data translation with rank-16 LoRA (Low-Rank Adaptation) produces a sharp task-type gradient: relational tasks gain measurably, while world-knowledge tasks regress. Bilingual Lexicon Induction aligns the French embeddings to GPT-2 at p@1 = $68.84 \pm 8.61$\%, $18\times$ above chance, suggesting cross-lingual alignment tracks acquired grammatical competence rather than training duration. 
An ablation study shows that single-token zero-shot scoring is dominated by tokenizer and template artifacts at the child scale, motivating tokenizer-swap sensitivity, placebo-controlled prompting, and native-language minimal-pair benchmarks as standard diagnostics.
\end{abstract}

\section{Introduction}
\label{sec:intro}

The BabyLM challenge \citep{warstadt2023babylm,hu2024babylm,charpentier2025babylm} asks whether language models can be trained on plausible amounts of input. 
Submissions in the Strict track are constrained to 100M words, and are evaluated on grammatical, world-knowledge, and downstream-task probes. 
This budget makes data curation and architectural choices first-order determinants of performance and turns child-scale pretraining into a tractable laboratory for both cognitive-science-inspired hypotheses about acquisition and engineering questions about efficiency.

Submissions through 2025 have all been in English, raising the question of whether the lessons drawn at the child scale are properties of language acquisition in general or of English specifically. The 2026 edition lifts this restriction \citep{babylm2026cfp}, and BabyBabelLM \citep{jumelet2026babelm} provides a multilingual training corpus across 15 languages. 
We submit to this edition MéTRON-FR, a 125M-parameter GPT-2 pretrained for five epochs on the French-only MéTRON-FR Strict corpus, with sentence-level oversampling driven by a Haitian-Creole lexical oracle. 
We evaluate on both the official BabyLM 2025 pipeline and with the Quebec French benchmarks \citep{beauchemin2025qfrblimp,beauchemin2025qfrcola}.

At the date of writing, MéTRON-FR's $85.97\%$ on QFrBLiMP exceeds the best native-BLiMP (Benchmark of Linguistic Minimal Pairs; \citealp{warstadt2020blimp}) score posted in any of the three official languages of the 2026 multilingual track: $80.75\%$ on English BLiMP (Strict, $\approx$100M words), $81.70\%$ on Dutch BLiMP-NL (50M words), and $78.60\%$ on Chinese ZhoBLiMP (50M words).\footnote{\href{https://huggingface.co/spaces/BabyLM-community/BabyLM-Leaderboard-2026}{Leaderboard snapshot 2026-05-06}} Each version of BLiMP is a separate monolingual native-language minimal-pair evaluation instrument.

Our single central message is that a French-only child-scale model attains strong \emph{native} grammatical competence, and that at this scale single-token zero-shot scoring is dominated by tokenizer and answer-template artifacts; everything else in the paper (cross-lingual transfer via BLI and CL GLUE, and the forensic ablations) is evidence in service of that message and of the reporting diagnostics we recommend for non-English child-scale submissions.
Our contributions are: \textbf{1)} We submit, to our knowledge, the first model trained from scratch under the BabyLM 100M-word constraint for evaluation on the Quebec dialect of French. \textbf{2)} We release a cross-lingual (CL) GLUE adaptation protocol that combines French task-data translation with rank-16 LoRA, and characterize a model's failure modes via three contrastive levers. \textbf{3)} We adapt Bilingual Lexicon Induction (BLI) as a child-scale geometric diagnostic of CL structural alignment.
The model, corpus, bilingual lemma bridge, and all evaluation scripts are available at \href{https://github.com/adamzwasserman/babylm}{\texttt{github.com/adamzwasserman/babylm}}.

The rest of this paper is organized as follows. In \autoref{sec:related}, we present a review of small-scale and multilingual LM work amongst native non-English minimal-pair benchmarks. \autoref{sec:approach} describes the model, corpus, and evaluation protocol. \autoref{sec:results} reports results on native Quebec French and on the official BabyLM suite. \autoref{sec:methodo} presents an ablation study isolating tokenizer-swap sensitivity, placebo-controlled prompting, and a lexical-frequency confound in the translated BLiMP Supplement, then draws methodological recommendations for child-scale non-English submissions (\autoref{sec:discussion}). We conclude the article in \autoref{sec:conclusion}.

\section{Related Work}
\label{sec:related}

\paragraph{Non-English and multilingual child-scale LMs.} Strict-track BabyLM entries from 2023 to 2025 \citep{warstadt2023babylm,hu2024babylm,charpentier2025babylm} are English-only and have advanced primarily through architectural innovation, with ELC-BERT \citep{charpentier2023elcbert} and GPT-BERT \citep{charpentier2024gptbert} as the leading lines of work. 
The 2026 multilingual axis \citep{babylm2026cfp} is enabled by BabyBabelLM \citep{jumelet2026babelm}, a developmentally plausible corpus across 15 languages, including French. 
Cross-linguistic studies with matched architectures document large differences in training efficiency across typologically diverse languages \citep{wasserman2026scaling}, with morphologically rich languages reaching grammatical competence with orders-of-magnitude fewer tokens than English.

\paragraph{Native non-English minimal-pair and acceptability benchmarks.} A French-trained model evaluated on translated BLiMP conflates model competence with translation quality and with lexical-frequency artifacts of the translation pipeline. 
Native targeted-syntactic and acceptability corpora that avoid this confound have been released for several languages following the methodology of \citet{warstadt2020blimp}; the Quebec-French Benchmark of Linguistic Minimal Pairs (QFrBLiMP; \citealp{beauchemin2025qfrblimp}) and the Quebec-French Corpus of Linguistic Acceptability (QFrCoLA; \citealp{beauchemin2025qfrcola}) provide this instrument for Quebec French.
Moreover, the COLE evaluation suite \citep{beauchemin2025cole} bundles these grammatical probes alongside lexical and cultural-knowledge tasks \citep{beauchemin2025qfrcore}.

\paragraph{CL transfer and structural alignment.} Two complementary methods inform our CL setup. 
BLI learns an orthogonal map between embedding spaces \citep{mikolov2013bli,conneau2018muse} and provides a geometric measurement of CL alignment that does not depend on translation. Yet, it was never adapted as a child-scale diagnostic to connect the resulting measurements to the formal-versus-functional linguistic competence distinction \citep{mahowald2024dissociating}. For downstream CL evaluation, parameter-efficient fine-tuning via Low-Rank Adaptation \citep{hu2021lora} preserves the pretrained base by construction and lets us run a four-lever experiment grid on a single submitted checkpoint without losing the grammatical signal measured natively.

\section{Methodology}
\label{sec:approach}

We address three orthogonal experimental questions: native grammatical competence (\autoref{sec:results-native}), CL transfer along four levers (BLI in \autoref{sec:results-bli}; GLUE adaptation in \autoref{sec:results-glue}), and forensic ablations isolating tokenizer, prompt-template, and lexical-frequency confounds in benchmark scoring (\autoref{sec:methodo}). All experiments share a single submitted checkpoint; no checkpoint is selected per experiment.

\subsection{Architecture and Training}
\label{sec:arch}

MéTRON-FR (Greek \emph{métron} for \enquote{measurement}) is a 125M-parameter GPT-2 \citep{radford2019gpt2} pretrained for five epochs on the MéTRON-FR Strict corpus (\autoref{sec:corpus} and \autoref{sec:oracle}). 
We select the best checkpoint as the empirically located grammatical-competence peak; intermediate checkpoints are released at the BabyLM-mandated cadence. Architecture hyperparameters, training schedule, hardware, GPU budget, and the released checkpoint manifest are detailed in \autoref{app:reproduction}.

\paragraph{Reproduction protocol.} The submitted checkpoint is the headline run; for variance estimation, we report a 5-seed reproduction ($[42\text{--}46]$) at matched corpus, tokenizer, and hyperparameters. Results in \autoref{sec:results} report mean $\pm$ standard deviation across the 5 seeds, alongside the submitted-checkpoint headline values (epoch 3, seed 42); the per-seed epoch-selection rule is given in \autoref{app:reproduction}.

\subsection{Tokenizer}
\label{sec:tokenizer}

We use a 50{,}000-token byte-pair encoding (BPE) \citep{sennrich2016bpe} tokenizer trained on the French training corpus itself (\autoref{app:tokenizer}). Three alternatives were compared in an ablation study: 16K/50K/64K child-directed speech-trained (CDS-trained). 
The 50K vocabulary yielded the lowest perplexity and the best downstream stability. 
The tokenizer is not a tuning parameter in our analysis. 
Indeed, \autoref{sec:tokenizer-swap} reports a forensic ablation in which only the tokenizer is swapped, with corpus, architecture, training steps, optimizer, and seed held constant.

\subsection{Corpus Composition}
\label{sec:corpus}

The training corpus is French-only and combines three sources, totalling 92{,}469{,}402 words (92.47M) as counted on the released \path{train_french.txt}.
A grammatically dense base draws on child-directed speech and accessible French text: $\approx\!2.10$M words from CHILDES French \citep{macwhinney2000childes}, plus $\approx\!24.90$M words from the non-subtitle portion of the French component of BabyBabelLM \citep{jumelet2026babelm}.
The remainder, $\approx\!66$M words, comes from the filtered French Wikipedia (see \autoref{app:reproduction} for details).
Per-source figures are counts of the source pools before oracle-weighted assembly (\autoref{sec:oracle}) and therefore do not sum exactly to the released total; the 92.47M figure is the whitespace word count of the released corpus file.
A hand-curated bilingual French--English lemma bridge is released alongside the corpus but is \textbf{not part of the training data}: its sole role is to supply the seed dictionary of the BLI diagnostic (\autoref{sec:results-bli}). It consists of 73 lemma entries (30 verbs, 18 nouns, 15 adjectives, 10 function words) with full inflection tables and English glosses, enumerating 715 distinct (FR, EN) form-level pairs. The corpus builder also exposes a variant harness that interleaves the bridge into the training text; the submitted model was trained on the French-only harness, and no bridge text enters it.
The corpus is pan-French rather than Quebec-specific: no Quebec-restricted corpus of comparable size exists at the BabyLM scale, and most QFrBLiMP phenomena (syntactic, morphological, agreement) are shared across French varieties; the gap with the Quebec normative prescription is isolated in the anglicism-related subscore and discussed in \autoref{sec:results-native}.
This simple recipe outperforms a more aggressively engineered reallocation with EWoK-targeted Wikipedia (EWoK, Elements of World Knowledge; \citealp{ivanova2025ewok}) and instruction blocks; see the ablation in \autoref{sec:tokenizer-swap}.

\subsection{The Haitian-Creole Lexical Oracle}
\label{sec:oracle}
Sentences are not sampled uniformly from the three sources. We oversample by the density of French lemmas whose cognates survived into Haitian Creole. Our working hypothesis, informed by creole-genesis research \citep{lefebvre1998creole,mufwene2001ecology}, is that because creoles form under acute communicative pressure, lemmas surviving this contact filter are plausibly disproportionately load-bearing for grammatical communication.
Concretely, we map a high-frequency Haitian-Creole (HC) lemma list to a French cognate set and oversample sentences dense in these lemmas; the extraction procedure, the cognate set, and the oversampling weight are detailed in \autoref{app:oracle}.
\textbf{No Haitian Creole text enters training}.
The lemmas act as an oracle lexical filter on French source material only. Using HC as an instrument rather than a training source is consistent with prior evidence that mixing analytical languages into morphologically rich corpora degrades the training dynamics of the rich language \citep{wasserman2026scaling}.
French is the morphologically richer member of the French--HC pair on every category that QFrBLiMP probes (verbal inflection, determiner-noun and subject-verb agreement); treating HC as a lexical filter rather than a training distribution preserves these French regularities while still drawing on the contact-filtered survival signal as a corpus-weighting heuristic.
We present the oracle as an \emph{exploratory} corpus-weighting heuristic, not as a validated contribution: its effect is not isolated against an unweighted baseline in this paper, and the clean-rebuild control (\autoref{sec:tokenizer-swap}) bounds any oracle-specific gain to within the training-time noise floor (\autoref{sec:limitations}). We report it because the recipe is what we submitted, and flag the unweighted-baseline ablation as follow-up work.

\subsection{Evaluation Suite}
\label{sec:eval-suite}

\paragraph{Official BabyLM 2026 leaderboard.} We report scores on the four instruments of the released 2025 evaluation pipeline. BLiMP \citep{warstadt2020blimp} is a zero-shot targeted-syntactic benchmark of English minimal pairs grouped into 67 paradigms (each a good/bad sentence pair), scored by comparing per-token log-probabilities. The BLiMP Supplement adds five pragmatic and discourse paradigms (e.g.\ question-answer congruence, subject-aux inversion) in the same format. (Super)GLUE \citep{wang2018glue} is a set of seven classification tasks evaluated after fine-tuning. EWoK \citep{ivanova2025ewok} probes world knowledge across physical, spatial, and social domains as a binary choice. The leaderboard reports both per-task accuracies and an aggregate score that averages across these instruments; a French-trained model is expected to be near chance on the English zero-shot probes, so we read the English-suite numbers as a lower-bound sanity check rather than a competitive target.

\paragraph{Native Quebec-French grammatical competence.}
Neither QFrBLiMP nor QFrCoLA is part of the official BabyLM suite; they are native Quebec-French instruments we adopt as our primary grammatical probe, on the argument (\autoref{sec:related}) that a native minimal-pair benchmark avoids the translation and lexical-frequency confounds of a translated one. They are complementary to, not a substitute for, MultiBLiMP \citep{jumelet2025multiblimp}, whose French portion targets the metropolitan standard: QFrBLiMP encodes the Office québécois de la langue fran\c{c}aise (OQLF) Quebec norm, so the two measure different prescriptive targets rather than the same one at different difficulty. We map the QFrBLiMP buckets to the translated-BLiMP paradigm breakdown per phenomenon in \autoref{app:blimp-paradigms}.
We use two native Quebec-French corpora. First, the QFrBLiMP \citep{beauchemin2025qfrblimp} a corpus of 1{,}761 minimal pairs annotated for 20 fine-grained linguistic phenomena of Quebec French, taken from the \emph{Banque de dépannage linguistique} (BDL), a public normative grammar resource maintained by the OQLF, and annotated by twelve native Quebec-French annotators. The 20 phenomena span syntactic, morphological, lexical-semantic, and lexical-normative judgments.
Second, the QFrCoLA \citep{beauchemin2025qfrcola} a sentence-level acceptability judgment corpus extracted from the BDL, with an additional out-of-domain split drawn from the \emph{Académie fran\c{c}aise} normative journal. Each sentence is labelled acceptable or unacceptable, where 69.50\% of sentences are labelled acceptable. The in-domain release totals 25{,}153 sentences, plus a separate 2{,}675-sentence out-of-domain set.


\paragraph{CL measurements.} Because English benchmarks cannot be applied directly to a French-trained model without translation or CL adaptation, we also report a structural CL measurement via BLI against frozen English embedding matrices (\autoref{sec:results-bli}) and a four-lever experiment grid for CL GLUE (\autoref{sec:results-glue}).

\subsection{Evaluation Metrics}
\label{sec:eval-metrics}

The minimal-pair benchmarks (QFrBLiMP, BLiMP, BLiMP Supplement, EWoK) are scored zero-shot as item-level accuracy per paradigm and overall; QFrCoLA is fine-tuned and reported as accuracy and Matthews Correlation Coefficient (MCC) \citep{matthews1975coefficient}; CL GLUE is fine-tuned with rank-16 Low-Rank Adaptation (LoRA) \citep{hu2021lora} on French-translated task data and reported as per-task accuracy and a five-task mean (\autoref{sec:results-glue}); and BLI reports the orthogonal Procrustes fit and word-translation precision@$k$ on a held-out split (\autoref{sec:results-bli}). The exact scoring rules, the MCC reference points, and the fine-tuning setup are detailed in \autoref{app:metrics}.

\section{Results}
\label{sec:results}

\subsection{Native Quebec-French Evaluation}
\label{sec:results-native}

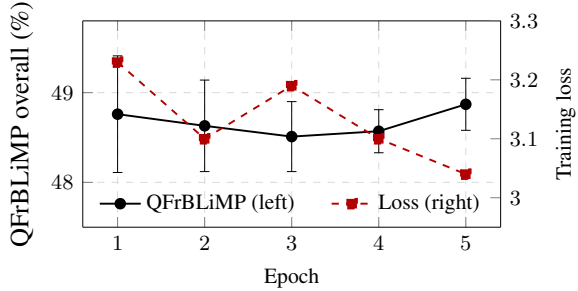
\begin{figure}[t]
    \centering
    \resizebox{\columnwidth}{!}{%
    \begin{tikzpicture}
        \begin{axis}[
            name=acc,
            width=\columnwidth,
            height=4.6cm,
            xlabel={Epoch},
            ylabel={QFrBLiMP overall (\%)},
            xmin=0.6, xmax=5.4,
            ymin=47.5, ymax=49.8,
            xtick={1,2,3,4,5},
            axis y line*=left,
            grid=major,
            grid style={dashed, gray!30},
            tick label style={font=\footnotesize},
            label style={font=\footnotesize},
            ylabel near ticks,
            legend style={
                at={(0.02,0.02)}, anchor=south west,
                font=\footnotesize,
                draw=none, fill=none,
                row sep=-1pt,
            },
        ]
            \addplot+[
                mark=*,
                mark options={fill=black},
                color=black,
                thick,
                error bars/.cd,
                y dir=both, y explicit,
                error bar style={black},
            ] coordinates {
                (1, 48.76) +- (0, 0.65)
                (2, 48.63) +- (0, 0.51)
                (3, 48.51) +- (0, 0.39)
                (4, 48.57) +- (0, 0.24)
                (5, 48.87) +- (0, 0.29)
            };
            \addlegendentry{QFrBLiMP (left)}
        \end{axis}
        \begin{axis}[
            width=\columnwidth,
            height=4.6cm,
            at=(acc.south west), anchor=south west,
            xmin=0.6, xmax=5.4,
            ymin=2.95, ymax=3.30,
            xtick=\empty,
            axis x line=none,
            axis y line*=right,
            ylabel={Training loss},
            ylabel near ticks,
            tick label style={font=\footnotesize},
            label style={font=\footnotesize},
            legend style={
                at={(0.98,0.02)}, anchor=south east,
                font=\footnotesize,
                draw=none, fill=none,
                row sep=-1pt,
            },
        ]
            \addplot+[
                mark=square*,
                mark options={fill=red!70!black},
                color=red!70!black,
                thick, dashed,
            ] coordinates {
                (1, 3.23)
                (2, 3.10)
                (3, 3.19)
                (4, 3.10)
                (5, 3.04)
            };
            \addlegendentry{Loss (right)}
        \end{axis}
    \end{tikzpicture}%
    }
    \caption{QFrBLiMP overall accuracy per epoch (mean $\pm$ std across 5 reproduction seeds; \autoref{sec:arch}) versus training loss on the submitted run. Grammatical accuracy stays within a $\sim\!0.40$pp band across the 5 epochs while training loss continues to decline, monotonically from epoch 3 onward, illustrating that grammatical competence saturates before perplexity does at this scale.}
    \label{fig:qfrblimp-trajectory}
    \vspace{-1.25em}
\end{figure}

\paragraph{QFrBLiMP.}
The submitted checkpoint reaches $85.97 \pm 0.17$\% overall on QFrBLiMP at epoch 3, the empirically located grammatical-competence peak (point estimate from the submitted run; variance from the 5-seed reproduction described in \autoref{sec:arch}). We report accuracy aggregated into four broader buckets in \autoref{tab:qfrblimp}; the epoch-1 reference checkpoint \texttt{chck\_92M}, used downstream for adaptation comparisons, scores 83.53\%. \autoref{fig:qfrblimp-trajectory} shows the multi-epoch trajectory across the 5-seed reproduction. Grammatical accuracy oscillates within a narrow band of approximately 0.40pp across epochs while training loss continues to decline, monotonically from epoch 3 onward, indicating, at the BabyLM scale, that grammatical competence saturates before perplexity saturates. 

\begin{table}[t]
\centering
\small
\begin{tabular}{lc}
\toprule
Bucket & Accuracy \\
\midrule
Syntactic & $89.74 \pm 1.32$\% \\
Semantic & $87.19 \pm 1.70$\% \\
Morphological & $85.47 \pm 1.02$\% \\
Anglicism-related & $80.15 \pm 1.54$\% \\
\midrule
\textbf{Overall (submitted)} & $\mathbf{85.97 \pm 0.17}$\% \\
\bottomrule
\end{tabular}%
\caption{QFrBLiMP results on the submitted checkpoint, zero-shot, using the aggregated four high-level buckets used in \citet{beauchemin2025qfrblimp}. Bucket point estimates are from the submitted run; standard deviations are from the 5-seed reproduction (\autoref{sec:arch}). The epoch-1 reference checkpoint scores 83.53\% overall.}
\label{tab:qfrblimp}
\vspace{-1em}
\end{table}

\paragraph{Anglicisms as a probe of normative Quebec French.}
On QFrBLiMP a higher score is better: it means the model assigns lower probability to the ungrammatical (here OQLF-proscribed) member of each minimal pair. The anglicism-related bucket is the lowest of the four (\autoref{tab:qfrblimp}) because sensitivity to the prescriptive norm is the hardest of these signals to acquire at the child scale, not because the model is failing at grammar in general.
The anglicism-related sub-score reflects two simultaneous mechanisms. The first is phonotactic and morphological internalization: anglicism candidates with non-French coda clusters or stress patterns receive lower probability under any model that has internalized French regularities, regardless of normative status. 
The second is sensitivity to the prescriptive norm encoded in the BDL itself: many lexical items in the underlying lexical-semantics phenomenon are phonotactically licit French formations (e.g. \emph{matcher}) but flagged as unacceptable in the BDL because the OQLF prescribes a French-source equivalent (\emph{correspondre}) as the standard. A high sub-score, therefore, reflects either internalization of the prescriptive distribution itself or of morphological cues correlated with it. Results on a metropolitan-French benchmark may differ on anglicism-sensitive items, where descriptive metropolitan usage is more permissive; the morphological and syntactic sub-scores are expected to generalize. \autoref{app:qcfr} details the OQLF normative framework.

\paragraph{QFrCoLA.}
On the in-domain test set, the fine-tuned classifier achieves $69.68 \pm 0.50$\% accuracy and an MCC of $0.13 \pm 0.03$ across the 5-seed reproduction. The reference upper bound is BERT-base on English CoLA at MCC $0.52$, achieved with $\approx\!110$M parameters but on the order of 3.30B pretraining tokens. At the $\approx\!1.3$ tokens-per-word rate of English BPE, that budget is $\approx\!2.5$B words, so ours is at the same parameter scale with $\approx\!27\times$ fewer words; tokens and words are not interchangeable units, and we state the comparison word-for-word. The $0.13$ MCC characterizes what acceptability-judgment competence is measurable at the BabyLM word budget on a native non-English benchmark: positive and above chance, but well below that of a model trained at the standard scale.

\subsection{Official BabyLM Suite}
\label{sec:results-official}

\paragraph{Translated BLiMP and Supplement.}
On translated BLiMP, the submitted checkpoint scores 76.28\%. We report this strictly as a cross-lingual transfer number: our French-trained model is scored on French translations of the originally English items, so it is not directly comparable to English-native models evaluated on the original English items. For context only, and not as a like-for-like ranking, OPT-125M \citep[a 125M-parameter open pretrained transformer language model;][]{zhang2022opt} scores 75.30\% and ELC-BERT \citep{charpentier2023elcbert} 85.30\% on the English items. See \autoref{app:blimp-paradigms} for the per-paradigm breakdown.
The translated BLiMP Supplement scores 34.00\%, below chance, reported as a methodological finding rather than a performance claim (\autoref{sec:blimp-sup}). 

\paragraph{GLUE.}
Under the submitted configuration, GLUE reaches a mean of $60.90$\% across the five translated tasks, on par with the $61.21$\% English-LoRA baseline mean (discussed in \autoref{sec:results-glue}). We translate the official English GLUE splits into French via an LLM pass, preserving the original alignment targets; the translation script is released with the submission. We treat this as a downstream adaptation choice rather than a pretraining data violation under the 2025 rule change: no new task content is introduced, and each French example is deterministically derivable from its English source.

\paragraph{EWoK.}
EWoK falls inside the 95\% chance interval $[47.90\%, 52.10\%]$ for $n\!=\!2200$ across all four interventions we tested (\autoref{sec:ewok}). We report it as a clean structural null result that bounds the task definition at the child scale.

\paragraph{Weighted leaderboard score.}
On the BabyLM-weighted metric, the submission scores 62.80\%, at or above all Strict-track reference points re-evaluated on the 2025 pipeline (GPT-BERT \citealp{charpentier2024gptbert} NLP 57.8--63.0, macro 40.8--43.5; GPT-2 naive; 2025 Strict-track winners; \citealp{charpentier2025babylm}; \autoref{app:strict-comparison}). We read this not as a competitive outperformance but as evidence that the language-choice axis is empirically larger than the architectural axis the field has been optimizing on.

\subsection{CL Structural Alignment via BLI}
\label{sec:results-bli}

The first CL question, i.e. \enquote{does a French-trained model exhibit structure recognizably aligned with that of English-trained models?}, admits a direct geometric answer via BLI \citep{mikolov2013bli,conneau2018muse}: we learn a frozen orthogonal map between the two models' embedding spaces on a seed dictionary, then measure held-out word-translation precision@$k$ by nearest-neighbour retrieval in the mapped space. The seed dictionary is the form-level expansion of the 73-entry bilingual lemma bridge of \autoref{sec:corpus}, a released resource that is not part of the training corpus; its 715 enumerated pairs reduce to 242 (FR, EN) pairs (194 to fit the map, 48 held out) after a morphology-aware parser pass; the closed-form Procrustes derivation and the parser-pass details are in \autoref{app:bli-method}. We compare the BabyLM French model against two English embedding targets (\autoref{tab:bli}): (a) GPT-2 \citep{radford2019gpt2}, 117M parameters trained on $\approx\!8$B WebText tokens and fully grammatically competent; (b) a matched-architecture 125M English model that plateaued at chance (40\%) on English grammar probes despite being trained on 6.50B C4 tokens \citep{wasserman2026scaling}.

\begin{table}[t]
\centering
\resizebox{\columnwidth}{!}{%
\begin{tabular}{lcccc}
\toprule
Target & Fit & p@1 (\%) & p@5 (\%) & p@10 (\%) \\
\midrule
GPT-2 (competent EN) & $\mathbf{0.40 \pm 0.03}$ & $\mathbf{68.84 \pm 8.61}$ & $\mathbf{91.87 \pm 6.07}$ & $\mathbf{98.56 \pm 1.97}$ \\
Matched-arch failed-EN & 0.51 & 25.00 & 52.10 & 66.70 \\
Random orthogonal (ref) & $-0.20$ & 11.50 & 26.90 & 53.80 \\
Chance ($1/n_{\text{test}}$) & --- & 3.80 & 19.20 & 38.50 \\
\bottomrule
\end{tabular}%
}
\caption{BLI Procrustes alignment from the BabyLM French model to two English embedding spaces. The GPT-2 row reports mean $\pm$ std across the 5-seed reproduction (\autoref{sec:arch}); the matched-arch failed-EN row is from the submitted run only. Fit $\;= 1 - \|E_{\mathrm{FR}}W - E_{\mathrm{EN}}\|_F^2 / \|E_{\mathrm{EN}}\|_F^2$ (1.00 perfect, 0.00 no alignment); p@k is held-out word-translation precision at the $k$ nearest neighbours under the learned map.}
\label{tab:bli}
\vspace{-1.25em}
\end{table}

Alignment to GPT-2 reaches p@1 of $68.84 \pm 8.61$\% on held-out word pairs, an order of magnitude above chance, purely via a frozen $768 \times 768$ orthogonal matrix; content-word alignments are rank-1 nearly universally (\emph{fuir}$\to$\emph{flee}, \emph{donner}$\to$\emph{give}, \emph{cassé}$\to$\emph{broke}, \emph{imaginer}$\to$\emph{imagine}, \emph{révéler}$\to$\emph{reveal}), and morphological features map structurally (\emph{écrivent}$\to$\emph{write}, \emph{révélé}$\to$\emph{revealed}). We present twenty retrievals with commentary in \autoref{app:bli}. The decisive contrast is the matched-architecture English model that failed to acquire English grammar: alignment drops to p@1 of 25.00\% despite training on 6.50B tokens ($71\times$ more than ours, and 50\% more than GPT-2 on a per-token basis). Structural alignment tracks acquired grammatical competence, not duration or architecture.

\subsection{CL GLUE Experiment Grid}
\label{sec:results-glue}

We test four LoRA-adapted levers for CL GLUE on each reproduction seed \citep{hu2021lora}, varying the task-data language, the LoRA rank, and the number of epochs; the key contrast is \texttt{D+C} (task data translated to French, rank-16 LoRA) against the English-LoRA \texttt{Baseline}. The full lever specification is in \autoref{app:glue-levers}. An inference-time vocabulary-axiom alternative (lever \texttt{B}) is reported separately in \autoref{sec:placebo}, and LoRA leaves the French base model bit-identical (\autoref{sec:lora-preservation}).
The translation lever (\texttt{D+C}) does not dominate uniformly across tasks but produces a sharp task-type gradient (\autoref{tab:glue-grid}): the relational tasks BoolQ ($+3.45$pp), RTE ($+5.13$pp), and MRPC ($+2.89$pp) gain measurably over the English-LoRA baseline, whereas the world-knowledge task MNLI regresses ($-11.10$pp) and the discourse-coreference task WSC is statistically inconclusive ($-1.92$pp with $\sigma\!=\!8.88$pp). The mean across the five tasks is therefore approximately flat ($-0.31$pp), but the per-task structure is the contribution: training-language adaptation helps where the task probes structural relations and not where it probes encyclopedic content or extended discourse.
We state this gradient as conditional on two controls. First, the per-task deltas of \texttt{D+C} over the \texttt{Baseline} conflate LoRA rank (8 vs 16) with the language of the task data (English vs French-translated); lever \texttt{C} (rank-16 English-LoRA, run on all five tasks across the 5 seeds) isolates the rank effect within a single language and shows no consistent gain over the rank-8 \texttt{Baseline}, so the structure in \texttt{D+C} is attributable to training-language adaptation rather than to rank. Second, we do not run the symmetric control, i.e.\ the same French-translated + rank-16 LoRA pipeline applied to an English-native child-scale model; without it we cannot fully exclude that the pipeline benefits any model regardless of pretraining language. We flag this control as follow-up work (\autoref{sec:limitations}).

\begin{table}[t]
    \centering
    \resizebox{\columnwidth}{!}{%
    \begin{tabular}{lccccc}
    \toprule
    Task & \texttt{Baseline} & \texttt{A} & \texttt{C} & \texttt{D+C} & $\Delta$ \\
    \midrule
    BoolQ  & $62.36 \pm 0.33$ & $63.10 \pm 0.72$ & $62.89 \pm 0.13$ & $\mathbf{65.81 \pm 0.51}$ & $+3.45$ \\
    RTE    & $53.14 \pm 2.65$ & $53.36 \pm 2.71$ & $52.85 \pm 1.69$ & $\mathbf{58.27 \pm 2.39}$ & $+5.13$ \\
    MRPC   & $68.77 \pm 0.93$ & $69.22 \pm 0.91$ & $68.58 \pm 0.84$ & $\mathbf{71.67 \pm 0.73}$ & $+2.89$ \\
    WSC    & $58.46 \pm 4.16$ & $\mathbf{59.42 \pm 2.19}$ & $58.85 \pm 4.05$ & $56.54 \pm 8.88$ & $-1.92$ \\
    MNLI   & $63.31 \pm 0.93$ & $\mathbf{64.47 \pm 0.74}$ & $64.29 \pm 0.58$ & $52.21 \pm 3.83$ & $-11.10$ \\
    \midrule
    \textbf{Mean} & 61.21 & 61.91 & 61.49 & 60.90 & $-0.31$ \\
    \bottomrule
    \end{tabular}%
    }
    \caption{CL GLUE grid (mean $\pm$ std across 5 seeds); best per task in bold. $\Delta$ is the gain of \texttt{D+C} over the \texttt{Baseline}. MultiRC and QQP held at baseline values for budget reasons. Inference-time vocabulary axioms (lever \texttt{B}) are placebo-tested separately in \autoref{sec:placebo}.}
    \label{tab:glue-grid}
    \vspace{-1.25em}
\end{table}

\paragraph{LoRA preserves the base by construction.}
\label{sec:lora-preservation}
Each per-task LoRA adapter applied to the epoch-1 checkpoint yields an identical 83.53\% on QFrBLiMP because the base parameters are bit-identical and the adapter weights do not enter the autoregressive head. Full fine-tuning on BoolQ drifts the QFrBLiMP score by $-0.17$pp to 83.36\%. We adopt LoRA as the headline adaptation method on the strength of this construction-time preservation guarantee, which we argue is methodologically preferable to post-hoc verification at child scale, where every grammatical-competence parameter is hard-won. Full ablation in \autoref{app:lora}.

\section{Methodological Findings and Recommendations}
\label{sec:methodo}

We report four ablations whose results we read as constraints on the interpretation of child-scale zero-shot scoring. None are pre-registered; they emerged during the preparation of submissions and are presented as forensic diagnostics rather than confirmatory tests. Together they support a single observation: at 125M parameters and 100M words, single-token log-probability scoring on short-sequence benchmarks responds principally to lexical-distributional properties of the tokenizer and the answer template, with task-level competence a secondary signal at best.

\subsection{Tokenizer-Swap on GLUE-Axiomatic}
\label{sec:tokenizer-swap}

We initially trained a substantially reallocated 93M-word corpus (\enquote{v2}) with 38M words of EWoK-targeted Wikipedia, 25M words of instruction data in GLUE format, and a 16K CDS-trained BPE tokenizer. The v2 model regressed by 7.70 percentage points on GLUE-axiomatic relative to the simpler v1 corpus (\autoref{tab:v3-ablation}, rows v1 and v2). To isolate which intervention caused the regression, we trained four additional models on the v1 corpus, varying one component at a time: \texttt{v3a} rebuilds v1 with the same 50K Wikipedia tokenizer; \texttt{v3b} replaces 15M words of Wikipedia with the format-targeted Opus instructions from v2; \texttt{v3c} replaces 38M words of Wikipedia with the EWoK-targeted Wikipedia from v2; \texttt{v3d} uses the v3a corpus but the 16K CDS-trained tokenizer from v2.

\begin{table}[t]
\centering
\resizebox{\columnwidth}{!}{%
\begin{tabular}{lccc}
\toprule
Variant & QFrBLiMP & EWoK & GLUE-ax \\
\midrule
v1 (\texttt{chck\_92M}) & 83.53 & 50.95 & 51.00 \\
v2 (full reallocation)  & 82.68 & 50.59 & 43.30 \\
\midrule
v3a (v1 recipe rebuild)         & 83.08 & 51.59 & 46.10 \\
v3b (v3a + 15M Opus)            & 83.30 & 50.50 & 50.70 \\
v3c (v3a + 38M EWoK Wiki)       & 83.08 & 52.32 & 50.70 \\
\texttt{v3d (v3a + 16K tokenizer)} & 82.68 & 51.36 & \textbf{43.50} \\
\bottomrule
\end{tabular}%
}
\caption{Single-intervention ablation isolating which v2 component caused the 7.70pp GLUE-axiomatic collapse. Only v3d, the tokenizer swap, reproduces the v2 floor.}
\label{tab:v3-ablation}
\vspace{-1.5em}
\end{table}

\noindent v3d alone reproduces the entire v2 GLUE-axiomatic collapse: 43.50\% versus v2's 43.30\%, with the same corpus as v3a, v3b, and v3c. The format-targeted instructions and the EWoK-targeted Wikipedia, when introduced individually on top of the v1 recipe, do not regress GLUE-axiomatic. The 16K CDS tokenizer swap, holding all other variables constant, does.

A second observation reinforces the reading. \texttt{v3a}, the clean rebuild of \texttt{v1} with a nominally identical recipe, scores 4.90pp lower than \texttt{v1} itself. The difference was accidental, not designed: a different pre-tokenization shuffle seed reordered the sentences of an otherwise identical corpus. It changes neither the source material nor the word budget, yet it produces GLUE-axiomatic swings comparable to differences between models (\autoref{app:tokenizer}, \path{rebuild_seed_sweep.json}). The score is a property of the tokenizer's vocabulary against the scoring protocol's answer-template tokens, not of task-level competence. The fine-tuned LoRA numbers we adopt as the headline GLUE results (\autoref{sec:results-glue}) replace single-token scoring with a learned decision boundary and are not subject to this artifact. Any method comparing log-probabilities of single-token answers across prompts is vulnerable, and a benchmark that does not report tokenizer-swap sensitivity cannot distinguish task-competence from tokenizer-vocabulary differences. Full v3 protocol and v2 details in \autoref{app:tokenizer} and \autoref{app:methodo}.

\subsection{Placebo Control on Dictionary-Axiom Prompting}
\label{sec:placebo}

An inference-time approach that prepends a French--English vocabulary axiom block to each English benchmark item produced apparent gains of 5--9 pp on three GLUE tasks (BoolQ, RTE, MNLI). A placebo control falsified the interpretation. We re-ran with three axiom variants under the same French answer template (\autoref{tab:placebo}):

\begin{table}[t]
\centering
\small
\begin{tabular}{lcccc}
\toprule
Task & Bare & Targeted & Placebo & Tautology \\
\midrule
BoolQ & 45.20 & 49.20 & 50.00 & 48.20 \\
RTE   & 54.00 & 54.00 & 54.00 & 54.00 \\
MNLI  & 32.20 & 35.20 & 35.20 & 34.00 \\
\midrule
Mean  & 43.80 & 46.10 & \textbf{46.40} & 45.40 \\
\bottomrule
\end{tabular}%
\caption{Dictionary-axioms placebo control. Targeted: relevant FR--EN pairs. Placebo: random, unrelated FR--EN pairs of the same number. Tautology: en=en pairs containing zero translation information.}
\label{tab:placebo}
\vspace{-1.5em}
\end{table}

\noindent Targeted minus bare is $+2.33$pp; placebo minus bare is $+2.60$pp; tautology minus bare is $+1.60$pp. The translation-specific effect (targeted minus placebo) is $-0.27$ pp: random, unrelated axioms produce as much improvement as targeted ones. The semantic content of the axiom block is irrelevant; the gain is structural prompting noise. The original 5--9pp effect was further inflated by an answer-template confound (English template without axioms vs.\ French template with axioms); with template held constant, the structural prompting effect is approximately 2pp. We retain this as a null result: structural prompting at the child scale yields a small, content-independent GLUE gain but does not constitute evidence of CL lexical transfer via vocabulary mappings.

\subsection{Translated BLiMP Supplement and EWoK at Child Scale}
\label{sec:blimp-sup}
\label{sec:ewok}

Two further results bound interpretation. The translated BLiMP Supplement score (34\%, \autoref{sec:results-official}) is below chance because the items contain a systematic lexical-frequency confound: in \emph{qa-congruence-easy}, for instance, the \enquote{good} French sentence consistently has a less-frequent content word (\emph{médecin} rather than \emph{chaise} as the subject of an embrassée-predicate) than the \enquote{bad} sentence, and the model prefers the higher-frequency alternative regardless of grammaticality. 
The score measures translation-specific lexical frequency in the training distribution, not the pragmatic inference phenomena that the English original probes; we report 34\% as honest evidence rather than as a competence claim. EWoK at $n\!=\!2200$ falls inside the 95\% chance interval $[47.90\%, 52.10\%]$ across all four interventions tested (raw pretraining, EWoK-targeted Wikipedia, dictionary axioms, retrieval-augmented prompting), with conditions ranging 49.86-50.95\%. The right tool for world knowledge at 125M / 100M words does not exist within the BabyLM envelope; inference-time tricks cannot substitute for parameters or data.

\subsection{Recommendations}
\label{sec:discussion}

The unifying observation across \autoref{sec:results} and \autoref{sec:methodo} is that no single instrument fits every evaluation decision at the child scale. \autoref{tab:right-tool} (\autoref{app:right-tool}) summarises the mapping for the seven decisions our pipeline made: each row is backed by an experiment in this paper, and every wrong-tool entry was tested and produced a null or negative result.

\paragraph{Native-language evaluation as the preferred axis.} The CL gap between QFrBLiMP (85.97\%, native) and translated BLiMP (76.28\%) is a 9.69-point gap on essentially the same kind of probe, differing only in whether the items are native or translated; translated minimal-pair benchmarks at child scale conflate model competence with translation quality and lexical-frequency artifacts (\autoref{sec:blimp-sup}). Where a native-language benchmark exists, it is the cleaner instrument; otherwise a placebo-controlled translation pass is a workable second-best.

\paragraph{Recommended reporting metrics.} For child-scale submissions scored by single-token log-probability on short-sequence benchmarks, two diagnostics should accompany headline accuracy: tokenizer-swap sensitivity (re-evaluating under a second BPE tokenizer trained on a different distribution) and placebo-controlled prompting (re-evaluating any prompt-prepended scoring under an unrelated-content prompt of matched length). Without them, a benchmark cannot separate task competence from tokenizer-vocabulary or prompt artifacts.

\section{Conclusion}
\label{sec:conclusion}

MéTRON-FR demonstrates that French-only child-scale pretraining produces a BabyLM submission scoring $85.97\%$ on QFrBLiMP, above the best native-BLiMP score in any other 2026 official language (\autoref{sec:intro}), and $62.80\%$ on the BabyLM-weighted metric, at or above the re-evaluated Strict-track reference points (\autoref{app:strict-comparison}), while exposing measurement gaps within the suite.

We do not argue that this submission supersedes the architectural line of work \citep{charpentier2023elcbert,charpentier2024gptbert}; we test a different axis, the choice of pretraining language, enabled by the 2025 rule change. Native-language minimal-pair benchmarks and protocol-level diagnostics belong in the standard reporting set for child-scale submissions relying on single-token zero-shot scoring.

\section*{Limitations}
\label{sec:limitations}

The submitted checkpoint is from a single pretraining seed; for variance estimation, we report a 5-seed reproduction (\autoref{sec:arch}) on the headline downstream evaluations (QFrBLiMP, QFrCoLA, BLI, CL GLUE). The 4.90 percentage-point gap on GLUE-axiomatic between two nominally identical recipe rebuilds (\autoref{sec:tokenizer-swap}, v1 vs.\ v3a) is a separate, training-time noise floor that bounds the resolution of any single-seed claim on this benchmark to approximately 5pp at this scale; we adopt fine-tuned LoRA numbers as the GLUE headline partly because they are not subject to this artifact.

Several quantitative sub-claims in this paper sit within or near that noise floor and should be read accordingly. The per-task gains in \autoref{tab:glue-grid} (BoolQ $+3.45$pp, RTE $+5.13$pp, MRPC $+2.89$pp; std $\le\!2.70$pp on the gain) are above the noise floor; the WSC and MNLI deltas are within or below it (WSC $\sigma\!=\!8.88$pp on \texttt{D+C}; MNLI std small but the regression magnitude is large), and the relative differences between v3a, v3b, and v3c in \autoref{tab:v3-ablation} are of comparable magnitude to the rebuild noise. The qualitative conclusions of those experiments, that the translated-data lever (\texttt{D+C}) helps short-input relational tasks while the inference-time-axiom lever (\texttt{B}) does not, and that only the tokenizer swap (\texttt{v3d}) reproduces the v2 GLUE-axiomatic floor, do not depend on point-estimate precision and survive within the noise floor; the per-task point estimates do.

The Quebec-French normative scope of QFrBLiMP and QFrCoLA is a deliberate choice but a real limitation. Anglicism-sensitive findings do not generalize to metropolitan French without re-validation, since the metropolitan norm is more permissive on lexical anglicisms (\autoref{app:qcfr}); morphological and syntactic findings are expected to generalize.

The single-language design cannot adjudicate whether the observed phenomena, namely the native-vs-translated minimal-pair gap, the BLI competence-vs-duration contrast, and the tokenizer and placebo sensitivities, are French-specific or general to child-scale non-English pretraining. Replicating the protocol across the BabyBabelLM languages \citep{jumelet2026babelm} with native minimal-pair instruments would test whether the right-tool mapping (\autoref{tab:right-tool}) holds cross-linguistically.

The methodological findings in \autoref{sec:methodo} are post hoc forensic diagnostics that emerged during the preparation of the submission, not pre-registered confirmatory tests; we report them as constraints on interpretation rather than as confirmatory claims. The Haitian-Creole lexical oracle is a corpus-curation heuristic whose contribution is not isolated against an unweighted baseline within this paper; the v3a clean-rebuild result (\autoref{sec:tokenizer-swap}) bounds any oracle-specific contribution to within the noise floor of the rebuild itself.

Our CL GLUE results (\autoref{sec:results-glue}) are not directly comparable to English-trained baselines such as ELC-BERT or GPT-BERT \citep{charpentier2023elcbert,charpentier2024gptbert}: those models report English-native fine-tuning on English task data, while ours reports rank-16 LoRA on French-translated task data. Two specific controls are missing from the present protocol. First, the per-task gains over the English-LoRA baseline conflate LoRA rank (8 vs 16) with the language of the task data (English vs French-translated); the rank-16 English-LoRA condition (\autoref{tab:glue-grid}, lever \texttt{C}, run on all five tasks in the 5-seed reproduction) does isolate the rank effect within the same language and shows no consistent gain over rank-8, supporting the reading that the per-task structure in lever \texttt{D+C} is driven by training-language adaptation rather than rank. Second, applying the same translated-French + rank-16 LoRA pipeline to an English-native child-scale model would test whether the protocol benefits any model regardless of pretraining language, or whether the gain is specific to having French-trained representations to adapt. This second control was outside the scope of computation for this submission and is flagged for camera-ready or follow-up work.

\section*{Ethical Considerations}
\label{sec:ethics}

The training corpus is assembled from public-domain or research-licensed French sources: CHILDES via \texttt{childes-db} \citep{macwhinney2000childes}, French Wikipedia, and the openly released \emph{babylm-fra} component of BabyBabelLM \citep{jumelet2026babelm}. The hand-curated bilingual lemma bridge is released as an evaluation resource for the BLI diagnostic and is not part of the training corpus. No human-subjects data is collected for this study, and no individually identifying content is introduced beyond what is already published in the source corpora; CHILDES transcripts are used within their documented research-release terms.

The Haitian-Creole lexical oracle uses a high-frequency word list as an instrument to oversample French sentences; no Haitian-Creole text enters training, and the procedure does not model, generate, or make claims about Haitian Creole as a language. We note the asymmetry: a creole derived from French serves here as an analytic tool for a French-only model, and we make no inference about Haitian Creole grammar or its speakers from the procedure.

QFrBLiMP encodes the prescriptive norm of the Office qu\'eb\'ecois de la langue fran\c{c}aise, an autonomous regulatory body whose recommendations differ from metropolitan French usage on the items the anglicism category probes (\autoref{app:qcfr}); submissions evaluated on QFrBLiMP should report this scope alongside the score.

The released model is a 125M-parameter French language model trained on 92.47M words. At this scale, generation quality is limited, and the model is not appropriate for user-facing applications; we release it as a research artifact for the BabyLM community, not as a deployment-ready system.

\section*{Use of AI Assistants}

Generative AI assistants (large language models) were used during the preparation of this submission. Their use was limited to: (i) copy-editing and language correction of author-written prose, including rephrasing for clarity and proofreading in the style of a Grammarly-type tool; and (ii) code-review and error-detection assistance applied to scripts that had been written and initially validated by the authors. All code in this submission was originally written by the authors; AI assistants were used to audit and review it, not to produce it. AI assistants were not used to generate research ideas, conduct experiments, run analyses, write substantive scientific argumentation, or perform literature search. The authors take full responsibility for all content.

\section*{Acknowledgements}
This research was made possible thanks to the support of a Canadian insurance company, NSERC research grant RDCPJ 537198-18 and an FRQNT doctoral research grant. We thank the reviewers for their comments regarding our work.

\bibliography{custom}

\appendix

\section{Right Tool, Right Job Summary}
\label{app:right-tool}

\begin{table}[h]
\centering
\footnotesize
\setlength{\tabcolsep}{4pt}
\begin{tabular}{@{}p{1.7cm}p{2.9cm}p{2.5cm}@{}}
\toprule
Job & Right tool & Wrong tool tried \\
\midrule
Corpus weighting & Cross-linguistic evolutionary filter (\autoref{sec:oracle}, exploratory) & Editorial intuition \\
Grammatical competence & French pretraining & English at child scale \\
Acceptability judgment & Fine-tune classification head (\autoref{sec:results-native}) & Zero-shot perplexity \\
CL GLUE & LoRA on translated task data (\autoref{sec:results-glue}) & Inference-time vocabulary axioms (\autoref{sec:placebo}) \\
CL structure & Procrustes on frozen embeddings (\autoref{sec:results-bli}) & Direct task transfer \\
World knowledge & More parameters or data (out of constraint) & Any child-scale intervention (\autoref{sec:ewok}) \\
Adaptation preservation & LoRA (\autoref{sec:lora-preservation}) & Full fine-tuning \\
\bottomrule
\end{tabular}
\caption{Right tool, right job: each row is backed by an experiment in this paper; every wrong-tool entry was tested and produced a null or negative result.}
\label{tab:right-tool}
\end{table}

\section{Evaluation Metric Details}
\label{app:metrics}

For QFrBLiMP, BLiMP, BLiMP Supplement, and EWoK, we score zero-shot by computing the per-token loss on each candidate sentence and selecting the lower-loss alternative, and report item-level accuracy aggregated per paradigm (i.e.\ per fine-grained linguistic phenomenon, e.g.\ \emph{determiner-noun agreement}) and across the benchmark. For QFrCoLA, we fine-tune a classification head on the in-domain train set and report accuracy and the Matthews Correlation Coefficient (MCC) \citep{matthews1975coefficient}; chance MCC is 0.00, and a fine-tuned BERT-base on English CoLA reaches MCC $0.52$ at substantially larger pretraining scale, which we use as an upper-bound reference. For CL GLUE, we fine-tune using rank-16 LoRA \citep{hu2021lora} on French-translated task data and report per-task accuracy, averaged across the five tasks (\autoref{sec:results-glue}). For BLI, we report the orthogonal Procrustes alignment fit and word-translation precision@$k$ on a held-out seed-dictionary split (\autoref{sec:results-bli}; \autoref{app:bli-method}).

\section{Haitian-Creole Oracle Procedure}
\label{app:oracle}

We extract the top 300 high-frequency Haitian-Creole (HC) lemmas and map them to a 333-lemma French cognate set with multi-form expansions such as \emph{nou} $\to$ \emph{nous} / \emph{on} / \emph{notre} / \emph{nos}. Each French training sentence is scored by target-lemma density, and high-density sentences are oversampled at weight 3.00 against a unit-weight default (\autoref{sec:oracle}). No Haitian-Creole text enters training; the lemmas act as a lexical filter on French source material only.

\section{CL GLUE Lever Specification}
\label{app:glue-levers}

The four LoRA levers of \autoref{sec:results-glue} (\autoref{tab:glue-grid}) are: 1) \texttt{Baseline} (rank-8 English LoRA, 3 epochs); 2) \texttt{A} (baseline + 5 epochs); 3) \texttt{C} (tuned rank-16 LoRA on English task data, $\alpha\!=\!32$); 4) \texttt{D+C} (\enquote{E3}: GLUE train/eval translated to French, then rank-16 LoRA). Lever \texttt{B} (inference-time vocabulary axioms) is placebo-tested in \autoref{sec:placebo}.

\section{Reproduction Checklist}
\label{app:reproduction}

\paragraph{Per-seed epoch selection.} Each of the 5 reproduction seeds ($[42\text{--}46]$) is trained for 5 epochs and evaluated at each epoch on QFrBLiMP; the per-seed best epoch is selected by argmax of QFrBLiMP overall accuracy and used as input to the downstream evaluations. The submitted headline checkpoint is epoch 3, seed 42.

\autoref{tab:repro-spec} summarises the model and training specification; \autoref{tab:repro-scripts} lists the scripts and output paths for each experiment reported in the main body. All scripts assume the submission directory as the working directory and are invoked via \texttt{uv run python}. Corpus word counts are whitespace-counted per the Strict-track convention, after deduplication.
The French Wikipedia component (\autoref{sec:corpus}) is filtered to exclude pure list pages, infobox text, and reference dumps.

\begin{table}[h]
\centering
\resizebox{\columnwidth}{!}{%
\begin{tabular}{ll}
\toprule
Component & Specification \\
\midrule
Architecture & 125M GPT-2 \\
Layers / dim / heads & 12 / 768 / 12 \\
Batch / sequence & 32 / 512 \\
Optimizer & AdamW (defaults) \\
Learning rate & $5\!\times\!10^{-4}$ \\
LR warmup & 1{,}000 steps \\
Epochs & 5 \\
Hardware & single RTX 4090 \\
Time / cost per epoch & $\approx\!24$ min / $\approx\!\$0.25$ \\
Total project cost & $\approx\!\$65$ \\
Tokenizer & 50K BPE, FR training corpus \\
Submitted checkpoint & epoch 3 (of 1--5) \\
\bottomrule
\end{tabular}%
}
\caption{Model and training specification.}
\label{tab:repro-spec}
\end{table}

\begin{table*}[h]
\centering
\begin{tabular}{>{\raggedright\arraybackslash}p{2.8cm}>{\raggedright\arraybackslash}p{5.7cm}>{\raggedright\arraybackslash}p{5.3cm}}
\toprule
Experiment & Scripts & Outputs / notes \\
\midrule
Pretraining (\autoref{sec:approach}) & \texttt{build\_french\_corpus.py}, \texttt{train.py} & \texttt{model/chck\_92M\_epoch3/} \\
QFrBLiMP (\autoref{sec:results-native}) & \texttt{eval\_qfrblimp.py} & 1{,}761 minimal pairs \\
QFrCoLA (\autoref{sec:results-native}) & \texttt{finetune\_qfrcola.py} & 3 epochs, learning rate $2\!\times\!10^{-5}$ \\
Translated BLiMP / Suppl. (\autoref{sec:results-official}) & \texttt{eval\_blimp\_translated.py} & translations under \texttt{fast\_eval\_improved/} \\
BLI (\autoref{sec:results-bli}) & \texttt{bli\_procrustes.py} (GPT-2); \newline \texttt{bli\_procrustes\_fractal.py} (failed-EN) & 242 FR--EN pairs (194 train, 48 held out) \\
CL GLUE (\autoref{sec:results-glue}) & \texttt{translate\_glue\_fr.py}; \newline \texttt{finetune\_glue\_lora\_fr\_tuned.py} & 7 per-task LoRA adapters, rank 16, $\alpha\!=\!32$, 3 epochs \\
Tokenizer-swap v3 (\autoref{sec:tokenizer-swap}) & \texttt{train\_tokenizer\_exp.py}, \texttt{build\_v3\_corpora.py} & logs under \texttt{full\_v3\_ablation/} \\
Dict-axioms placebo (\autoref{sec:placebo}) & \texttt{eval\_dict\_axioms\_*.py} & logs under \texttt{dict\_axioms/} \\
\bottomrule
\end{tabular}
\caption{Scripts and outputs for each experiment.}
\label{tab:repro-scripts}
\end{table*}

\section{Per-paradigm Translated BLiMP}
\label{app:blimp-paradigms}

\autoref{tab:blimp-paradigms} reports the 38-paradigm breakdown for the translated BLiMP evaluation of MéTRON-FR at epoch 3 (\autoref{sec:results-official}). Each paradigm has 200 unique minimal pairs; chance is 50\%. Fifteen paradigms score above 80\%; the lowest-scoring paradigms cluster in Principle A (anaphor binding) and gender-agreement categories, where French--English grammatical differences make the translation lossiest.

\begin{table*}[h]
\centering
\begin{tabular}{@{}>{\raggedright\arraybackslash}p{9cm}c@{}}
\toprule
Paradigm & Accuracy \\
\midrule
anaphor gender agreement & 43.50\% \\
anaphor number agreement & 74.00\% \\
animate subject transitive & 78.00\% \\
coordinate structure (complex left branch) & 87.50\% \\
coordinate structure (object extraction) & 56.50\% \\
determiner-noun agreement 1 & 93.50\% \\
determiner-noun agreement 2 & 100.00\% \\
determiner-noun agreement (irregular 1) & 85.00\% \\
determiner-noun agreement with adjective 2 & 98.50\% \\
distractor agreement (relative clause) & 78.50\% \\
ellipsis n-bar 1 & 77.00\% \\
expletive it, object raising & 72.50\% \\
inchoative & 77.50\% \\
intransitive & 74.00\% \\
irregular past participle adjectives & 82.50\% \\
irregular past participle verbs & 76.50\% \\
irregular plural subject-verb agreement 1 & 86.50\% \\
irregular plural subject-verb agreement 2 & 86.50\% \\
left-branch island (simple question) & 92.00\% \\
matrix question negative polarity item licensor present & 80.00\% \\
negative polarity item present 1 & 78.00\% \\
only-negative polarity item scope & 98.50\% \\
passive 1 & 73.00\% \\
principle A (case 1) & 43.00\% \\
principle A (case 2) & 82.00\% \\
principle A (domain 3) & 48.50\% \\
principle A (reconstruction) & 51.50\% \\
sentential negation negative polarity item licensor present & 100.00\% \\
sentential subject island & 69.00\% \\
superlative quantifiers 2 & 86.00\% \\
tough-vs-raising 1 & 65.00\% \\
tough-vs-raising 2 & 55.00\% \\
transitive & 67.50\% \\
wh-island & 92.00\% \\
wh-question subject gap & 74.50\% \\
wh-question subject gap (long distance) & 93.00\% \\
wh-vs-that, no gap, long distance & 80.50\% \\
wh-vs-that, gap, long distance & 41.50\% \\
\midrule
Overall & 76.30\% \\
\bottomrule
\end{tabular}
\caption{Per-paradigm translated BLiMP accuracy for MéTRON-FR at epoch 3.}
\label{tab:blimp-paradigms}
\end{table*}

\section{Strict-track Reference Comparison}
\label{app:strict-comparison}

\autoref{tab:strict-comparison} reproduces the Strict-track reference points cited in \autoref{sec:results-official} under the 2025 evaluation pipeline, drawn from \citet[Table~3]{charpentier2025babylm}.

\begin{table}[h]
\centering
\small
\setlength{\tabcolsep}{4pt}
\begin{tabular}{@{}lcc@{}}
\toprule
Model (Strict track) & NLP & Macro \\
\midrule
GPT-BERT, 7 variants & 57.8--63.0 & 40.8--43.5 \\
GPT-2 naive baseline & 55.4 & 36.2 \\
CLASS-IT (2025 HL win.) & 52.9 & 36.6 \\
Simple-Diffusion (2025 NLP win.) & 58.4 & 35.5 \\
\midrule
\textbf{MéTRON-FR (2026)} & \multicolumn{2}{c}{\textbf{62.80 (weighted)}} \\
\bottomrule
\end{tabular}
\caption{Strict-track reference scores under the 2025 evaluation pipeline (from \citealp{charpentier2025babylm}, Table~3): NLP-task average (GLUE-style fine-tuning) and macro average. GPT-BERT \citep{charpentier2024gptbert} is the prior challenge winner re-evaluated as the 2025 baseline; the seven variants combine \{causal, mixed, masked\} training objectives with \{MNTP, AR\} evaluation styles. MéTRON-FR's BabyLM-weighted score is a single composite reported by the same pipeline.}
\label{tab:strict-comparison}
\end{table}

\section{BLI Method Details}
\label{app:bli-method}

Given two frozen embedding matrices $E_{\mathrm{FR}}$ and $E_{\mathrm{EN}}$ and a seed dictionary, we learn an orthogonal transformation $W \in \mathbb{R}^{768 \times 768}$ minimizing $\|E_{\mathrm{FR}} W - E_{\mathrm{EN}}\|_F$ via closed-form Procrustes, and evaluate word-translation precision@$k$ by cosine nearest-neighbour retrieval in the mapped space (\autoref{sec:results-bli}). The reported fit is $1 - \|E_{\mathrm{FR}}W - E_{\mathrm{EN}}\|_F^2 / \|E_{\mathrm{EN}}\|_F^2$ (1.00 perfect, 0.00 no alignment).

The seed dictionary starts from the 715 form-level (FR, EN) pairs enumerated from the 73-entry bilingual lemma bridge of \autoref{sec:corpus}, a resource released with the submission that is not part of the training corpus. The \emph{language-aware parser pass} is a morphology-aware normalization step: it lemmatizes and, where needed, segments each surface form on both the French and English sides so that inflected variants map onto their base form, drops pairs that survive in only one vocabulary, and deduplicates the result to the (FR, EN) pairs that are single-token in both embedding matrices (a requirement of the closed-form Procrustes map). After this pass, 242 unique pairs are retained; 194 are used to fit $W$ and 48 are held out for precision@$k$ evaluation.

\section{BLI Alignment Qualitative Examples}
\label{app:bli}

\autoref{tab:bli-retrievals} shows twenty representative held-out French-to-English word-translation retrievals from the BLI experiment (\autoref{sec:results-bli}), under the learned orthogonal map $W$ aligning MéTRON-FR's embedding matrix to GPT-2's. \emph{Rank} is the position of the gold English translation in the top-$k$ cosine-similarity nearest neighbours. Content-word mappings reach rank 1 in most cases; function-word mappings and morphologically complex inflections frequently succeed at rank 2--5.

\begin{table*}[h]
\centering
\begin{tabular}{l l c p{10cm}}
\toprule
FR held-out & EN gold & Rank & Top-5 retrieved \\
\midrule
\emph{écrit} & writes & 2 & write, writes, read, articles, notice \\
\emph{l'} & the & 1 & the, the, a, this, in \\
\emph{allé} & gone & 1 & gone, watched, forgotten, scared, do \\
\emph{nous} & us & 2 & we, us, your, the, the \\
\emph{ce} & this & 1 & this, the, the, a, what \\
\emph{fuir} & flee & 1 & flee, speak, scared, conceal, gone \\
\emph{imaginer} & imagine & 1 & imagine, write, give, see, reveal \\
\emph{par} & by & 5 & in, the, a, the, by \\
\emph{cassé} & broke & 1 & broke, gone, forgotten, scared, revealed \\
\emph{écris} & write & 1 & write, flee, watch, so, read \\
\emph{dans} & in & 1 & in, the, the, a, by \\
\emph{révéler} & reveal & 1 & reveal, revealed, give, conceal, see \\
\emph{regarder} & watch & 3 & do, see, watch, give, watched \\
\emph{nous} & we & 1 & we, us, your, the, the \\
\emph{une} & a & 1 & a, the, the, this, in \\
\emph{donc} & so & 4 & the, the, a, so, this \\
\emph{donner} & give & 1 & give, do, see, speak, reveal \\
\emph{quoi} & what & 2 & what, what, your, a, we \\
\emph{caché} & concealed & 8 & watched, scared, revealed, gone, watch \\
\emph{révélé} & revealed & 1 & revealed, reveal, forgotten, watched, investigating \\
\bottomrule
\end{tabular}
\caption{Sample BLI retrievals (MéTRON-FR $\to$ GPT-2 via learned orthogonal map). Content verbs (\emph{fuir}, \emph{cassé}, \emph{donner}, \emph{révéler}) and morphologically inflected past participles (\emph{allé}, \emph{révélé}) align at rank 1. Function words (\emph{par}, \emph{donc}) are retrieved at slightly lower ranks but within the top 5.}
\label{tab:bli-retrievals}
\end{table*}

\section{Tokenizer Ablation Full Protocol}
\label{app:tokenizer}

The v3a--v3d ablation (\autoref{sec:tokenizer-swap}, \autoref{tab:v3-ablation}) was run with the same architecture, batch size, sequence length, optimizer, learning-rate schedule, and random seed (42) as the v1 submission. The four v3 models were trained for one epoch each on their respective corpora; QFrBLiMP is reported on the resulting \texttt{chck\_92M}-equivalent checkpoints. Tokenizer training corpora and seeds:

\begin{itemize}
\setlength{\itemsep}{0pt}
\item \texttt{50K-FR BPE} (v1, v3a, v3b, v3c): trained on the full v1 training corpus (\path{corpus/final/train_french.txt}, 92.47M words), vocab size 50{,}000 including the two special tokens \texttt{<|endoftext|>} and \texttt{<|padding|>}, byte-level pre-tokenizer with \texttt{add\_prefix\_space=False} and byte-level decoder, and no Unicode normalization step. BPE training is deterministic given the corpus and the vocabulary size, so no tokenizer seed is involved.
\item \texttt{16K-CDS BPE} (v2, v3d): trained on the 25M-word CHILDES + babylm-fra component of the v1 corpus, vocab size 16{,}000, otherwise identical to the \texttt{50K-FR} configuration.
\end{itemize}

\noindent The GLUE-axiomatic evaluation set is the inference-time axioms protocol of Lever B (\autoref{sec:results-glue}) on the five translated GLUE tasks, scored using the lowest-loss-token-among-the-template-options heuristic. The full release includes \path{scripts/train_tokenizer_exp.py}, the four trained tokenizer files, the v3a--v3d corpus chunks, and the per-task scoring logs at \path{submission/eval_logs/full_v3_ablation/}. The 4.90pp v1-vs-v3a rebuild noise (a positive control: same recipe, different seeds in pre-tokenization shuffle) is documented in the same log directory under \path{rebuild_seed_sweep.json}.

\section{Quebec French Normative Framework}
\label{app:qcfr}

QFrBLiMP encodes the prescriptive norm of the \textit{Office québécois de la langue fran\c{c}aise} (OQLF), an autonomous regulatory body created under the \emph{Charte de la langue fran\c{c}aise} (Loi 101, 1977; expanded by Loi 96, 2022). The OQLF defines normative Quebec French through the \emph{Grand dictionnaire terminologique} and the \emph{Banque de dépannage linguistique}; its mandate is both administrative and descriptive; its recommendations are binding for Quebec public administration, education, and commercial signage. The norm, therefore, has political constitution, not merely lexicographic authority.

The metropolitan French standard, by contrast, is descriptive: the\textit{ Académie fran\c{c}aise} issues recommendations without administrative force, and the metropolitan written-press and dictionary tradition is more permissive on lexical anglicisms in an informal register. The two norms diverge most sharply on the items the QFrBLiMP anglicism category probes:

\begin{itemize}
\setlength{\itemsep}{0pt}
\item \emph{checker} (QC anglicism) vs \emph{vérifier} (OQLF normative). Metropolitan informal usage admits \emph{checker} freely; metropolitan formal register prefers \emph{vérifier}.
\item \emph{matcher} (anglicism) vs \emph{correspondre} / \emph{aller avec} (OQLF normative). Metropolitan informal usage admits \emph{matcher} routinely.
\item \emph{fun} (anglicism) vs \emph{plaisir} / \emph{amusant} (OQLF normative). Metropolitan informal usage admits \emph{fun} as a noun and adjective.
\item \emph{cool} (anglicism) vs \emph{bien} / \emph{super} (OQLF normative). Metropolitan usage admits \emph{cool} across registers.
\item \emph{courriel} (OQLF coinage, normative QC) vs \emph{email}/\emph{e-mail} (metropolitan accepts; OQLF prefers \emph{courriel}).
\end{itemize}

A model evaluated on QFrBLiMP is being asked whether it has internalized the OQLF norm, not French grammaticality in some language-neutral sense. The 80.15\% anglicism sub-score (\autoref{tab:qfrblimp}) reflects two simultaneous mechanisms: (i) phonotactic and morphological internalization of French regularities, which assigns lower probability to anglicisms with non-French coda clusters or stress patterns regardless of normative status, and (ii) sensitivity to the OQLF norm itself, which flags phonotactically licit anglicisms as ungrammatical. The benchmark's strength is a well-defined ground truth backed by a regulatory body with continuous publication; its scope limitation is that performance does not translate one-to-one to a metropolitan-French informal-register benchmark.

\section{LoRA Preservation Details}
\label{app:lora}

\autoref{tab:lora-preservation} reports QFrBLiMP scores after each per-task LoRA adapter is applied to \texttt{chck\_92M} (epoch 1, baseline value 83.53\%). Every LoRA adapter produces an identical 83.53\% because the base model's parameters are bit-identical by construction and the adapter weights do not enter the autoregressive language-modelling head used for QFrBLiMP scoring. The submitted epoch-3 checkpoint exhibits identical preservation properties at its own QFrBLiMP baseline of 85.97\%. Full fine-tuning on BoolQ drifts the QFrBLiMP score by $-0.17$pp, indicating mild catastrophic forgetting at this scale.

\begin{table*}[h]
\centering
\begin{tabular}{lcc}
\toprule
Configuration & QFrBLiMP & GLUE (E3 best) \\
\midrule
\texttt{chck\_92M} (no FT)         & 83.53 & --- \\
+ E3 BoolQ adapter                 & 83.53 & 68.26 \\
+ E3 MNLI adapter                  & 83.53 & 53.79 \\
+ E3 MRPC adapter                  & 83.53 & 72.55 \\
+ LoRA MultiRC adapter             & 83.53 & 57.55 \\
+ LoRA QQP adapter                 & 83.53 & 73.14 \\
+ E3 RTE adapter                   & 83.53 & 61.15 \\
+ E3 WSC adapter                   & 83.53 & 73.08 \\
\midrule
Full FT on BoolQ                   & 83.36 & 66.20 \\
\bottomrule
\end{tabular}
\caption{QFrBLiMP after downstream LoRA fine-tuning. Every adapter produces the same 83.53\% because the base parameters are unchanged, and adapter weights do not enter the autoregressive head. Full FT drifts QFrBLiMP by 0.17pp.}
\label{tab:lora-preservation}
\end{table*}

The result is a parameter-efficient instance of the lottery-ticket hypothesis \citep{frankle2018lottery} at the adaptation layer rather than the pretraining layer: 0.65\% of total parameters (the rank-8 baseline LoRA) or 1.29\% (the rank-16 tuned LoRA) suffices to encode per-task downstream behaviour, while the remaining $\geq\!98.70\%$ serves as a fixed substrate. The construction-time guarantee replaces the empirical claim \enquote{we measured QFrBLiMP after fine-tuning, and it stayed high} with a structural one: the base parameters are bit-identical, so the QFrBLiMP score is unchanged by definition.

\section{Additional Methodological Details}
\label{app:methodo}

\paragraph{v2 corpus reallocation (\autoref{sec:tokenizer-swap}).} The v2 model was trained on a 93M-word reallocation: 25M words of grammatical base content (CHILDES + babylm-fra, identical to v1), 38M words of EWoK-targeted Wikipedia (filtered for vocabulary relevant to the physical, social, and spatial domains targeted by EWoK), 15M words of GLUE-format Opus instructions in French, 10M words of supplement-targeted Opus instructions, and 4.60M words of diverse instruction data. v2 regressed on every metric (\autoref{tab:v2-vs-v1}).

\begin{table*}[h]
\centering
\begin{tabular}{lccc}
\toprule
Metric & v1 & v2 & $\Delta$ \\
\midrule
QFrBLiMP & 83.53 & 82.68 & $-0.85$ \\
GLUE-axiomatic & 51.00 & 43.30 & $-7.70$ \\
EWoK & 50.95 & 50.59 & $-0.36$ \\
\bottomrule
\end{tabular}
\caption{v1 vs v2 corpus reallocation results. The 7.70pp GLUE-axiomatic regression motivated the v3 ablation (\autoref{sec:tokenizer-swap}).}
\label{tab:v2-vs-v1}
\end{table*}

\noindent A per-item diagnosis on the BoolQ validation set showed v2 predicting \enquote{non} (no) for 97.60\% of items, while v1 predicted \enquote{oui} (yes) for 66.60\% (gold yes-rate is 62.20\%). The v2 regression is therefore a token-calibration collapse rather than a task-level capability loss: the format-targeted instruction data shifted the output distribution toward \enquote{non}, and the resulting accuracy regression is the mechanical consequence of a benchmark whose gold distribution favours \enquote{yes}. The single-intervention v3 ablation (\autoref{sec:tokenizer-swap}) localized this calibration shift to the 16K CDS-trained tokenizer.

\paragraph{EWoK by-domain breakdown.} \autoref{tab:ewok-domain} shows the four interventions across the eleven EWoK domains. No domain shows statistically significant separation from chance for any intervention.

\begin{table*}[h]
\centering
\begin{tabular}{lcc}
\toprule
Intervention & EWoK & 95\% CI \\
\midrule
v1 raw pretraining & 50.95 & [47.90, 52.10] \\
v2 EWoK-targeted Wiki & 50.59 & [47.90, 52.10] \\
Dict-axioms (small bridge) & 49.86 & [47.90, 52.10] \\
Dict-axioms (big, 452 entries) & 50.59 & [47.90, 52.10] \\
Retrieval-augmented (top-3) & 50.95 & [47.90, 52.10] \\
\midrule
Chance (binary) & 50.00 & --- \\
\bottomrule
\end{tabular}
\caption{EWoK across interventions. All five conditions fall inside the 95\% chance interval $[47.90\%, 52.10\%]$ for $n\!=\!2200$.}
\label{tab:ewok-domain}
\end{table*}

\paragraph{Dict-axioms placebo answer-template confound.} The original 5--9pp axiom effect (\autoref{sec:placebo}) compared an English answer template (no axioms) against a French answer template (with axioms), conflating the template change with axiom introduction. With the template held constant, the structural prompting effect (targeted minus bare under the matched French template) is approximately 2 pp; the translation-specific effect (targeted minus placebo) is $-0.27$ pp.

\end{document}